\documentclass[10pt,twocolumn]{article}

\usepackage{iccv}
\usepackage{times}
\usepackage{epsfig}
\usepackage{graphicx}
\usepackage{amsmath}
\usepackage{amssymb}
\usepackage{booktabs}
\usepackage{multirow}
\usepackage[accsupp]{axessibility}  

\usepackage[pagebackref=true,breaklinks=true,colorlinks,bookmarks=false]{hyperref}

\usepackage[capitalize]{cleveref}
\crefname{section}{Sec.}{Secs.}
\Crefname{section}{Section}{Sections}
\Crefname{table}{Table}{Tables}
\crefname{table}{Tab.}{Tabs.}

\newcommand{\mysubsubsection}[1]{\noindent {\bf #1}:}

\newcommand{\fg}[1]{\textcolor{black}{#1}}
\newcommand{\rz}[1]{\textcolor{black}{#1}}
\newcommand{\rzz}[1]{\textcolor{black}{#1}}
\newcommand{\rzzz}[1]{\textcolor{black}{#1}}

\newcommand{\pg}[1]{}

\iccvfinalcopy 

\def\iccvPaperID{5418} 

\ificcvfinal\fi

\begin{document}

\title{HAL3D: Hierarchical Active Learning for Fine-Grained 3D Part Labeling}

\author{Fenggen Yu$^{1,2}$\thanks{Work carried out during internship at Amazon.} \qquad Yiming Qian$^{1}$ \qquad Francisca Gil-Ureta$^{1}$ \\
\qquad Brian Jackson$^{1}$ \qquad Eric Bennett$^{1}$ \qquad Hao Zhang$^{1,2}$
\\
$^1$Amazon \qquad $^2$ Simon Fraser University}

\maketitle
\ificcvfinal\thispagestyle{empty}\fi

\begin{abstract}
We present the first {\em active learning\/} tool for {\em fine-grained\/} 3D part labeling, a problem which challenges even the most advanced deep learning (DL) methods due to the significant structural variations among the intricate parts. For the same reason, the necessary effort to annotate training data is tremendous, motivating approaches to minimize human involvement. Our labeling tool iteratively verifies or modifies part labels predicted by a deep neural network, with human feedback continually improving the network prediction. To effectively reduce human efforts, we develop two novel features in our tool, hierarchical and symmetry-aware active labeling. Our human-in-the-loop approach, coined HAL3D, achieves close to error-free fine-grained annotations on any test set with pre-defined hierarchical part labels, with 80\% time-saving over manual effort. We will release the finely labeled models to serve the community.
\end{abstract}

\section{Introduction}
\label{sec:intro}

\begin{figure}[t!]
     \centering
     \includegraphics[width=0.99\linewidth]{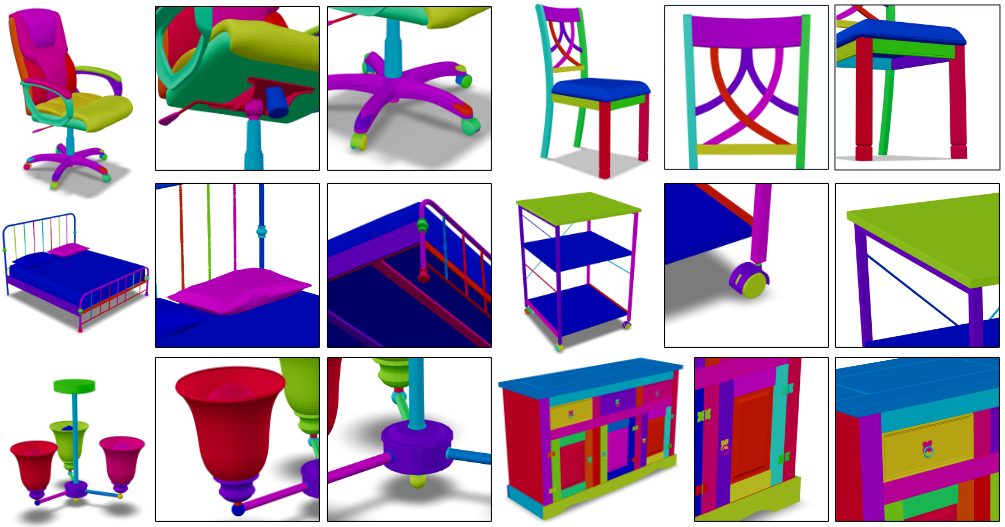}
   \caption{Fine-grained parts of the 3D models shown exhibit the kind of geometric and structural complexity and diversity that part labeling has to handle. No existing methods, whether learned or heuristic-based, could obtain close-to-fully-accurate labeling in these challenging cases, while our human-in-the-loop active learning tool can \rzz{approach full accuracy}, barring human errors.}
\label{fig:teaser}
\end{figure}

Semantic shape segmentation and labeling~\cite{DLSegSurvey,Shamir2008} is a classical problem with numerous applications~\cite{Mitra2013}, including shape/part recognition, retrieval, indexing, and attribute transfer.
We consider the problem of {\em fine-grained\/} 3D part labeling, a task that has not received as much attention as the typical semantic segmentation into coarse/major parts (e.g., backs, seats, and legs for chairs). While these coarse semantic groups support high-level visual perception, they do not possess sufficient granularity to address shape properties related to motion, function, interaction, or construction. Indeed, fine-grained object parts, e.g., the individual wheels and the small mechanical parts in a swivel leg, are both function-revealing and build-aware (see \cref{fig:teaser}), with strong ties to how the objects were physically assembled or virtually built by artists. Real-world applications such as product inspection, assembly, customization, and robot interaction all operate on fine-grained object parts.

Learning to label fine-grained parts is a highly challenging problem due to the great structural variations among the small and intricate parts, even for shapes that belong to the same category, as shown in \cref{fig:teaser}. 
On a challenging dataset with average part counts per category ranging from 15 to more than 500, state-of-the-art learned models such as~\cite{wang2021fgseg} can only achieve mean IoU scores below 50\%, out of a maximum of 100\% for ground truth (GT) results.
Also challenging is the tremendous data annotation effort needed to train methods to segment and label fine-grained parts. On the other hand, for applications which demand de facto \rzz{{\em full}} accuracy, e.g., for diagnoses in medicine and quality assurance in e-commerce, one can hardly rely on a fully automatic method to reach the mark. It is arguable that the only means to attain full accuracy is to have humans validate all the results, as \rzz{in the case of data annotation}.

\begin{figure*}[ht!]
\centering
\includegraphics[width=0.99\linewidth]{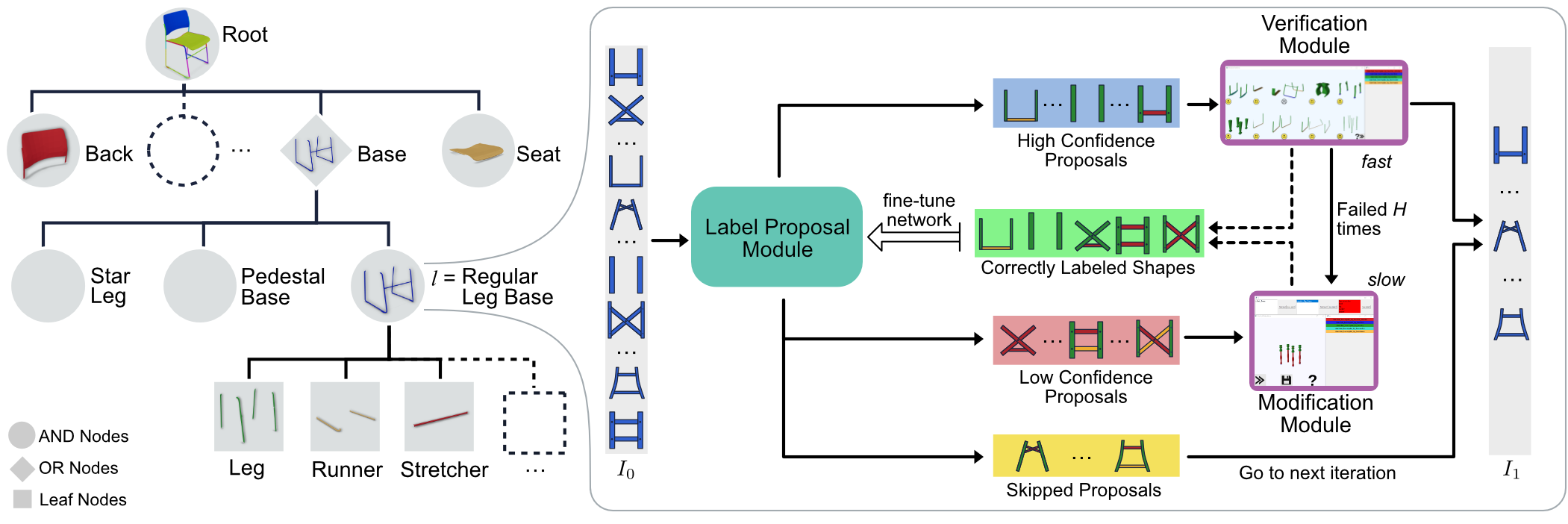}
\caption{Our human-in-the-loop, hierarchical active learning (HAL3D) tool for fine-grained 3D part labeling. 
   The input consists of a set of test shapes each pre-segmented into parts. The labeling proceeds hierarchically, following a tree structure (left) that organizes the hierarchical part labels, from coarse labels (top) to fine-grained labels (bottom). For a node in the label tree (with label $l$ = ``Regular Leg Base" in the illustration), the input $I_0$ is the subset of parts, from the entire set of shapes, that are labeled $l$ by its parent. When labeling parts with $l$, the label proposal module first assigns refined labels for each part. Then, proposals are sorted by \fg{the mean label probability over the parts of each shape}, with the high-confidence (HC) proposals passed to the verification step, which stops once the rate of failed shapes reaches a threshold. The low-confidence (LC) proposals are passed to the label modification module. Correctly labeled shapes fine-tune the network, while skipped and failed shapes, in set $I_1$, go to the next iteration for labeling. The iterations terminate when all shapes have passed human validation.}
\label{fig:pipeline}
\end{figure*}

In this paper, we present the first {\em active learning\/} tool for fine-grained 3D part labeling. Given a set of 3D shapes segmented into fine parts, our labeling tool assigns one of the predefined labels to each part. These input parts are deemed {\em atomic\/} (i.e., indivisible); they can be as low-level as mesh triangles and as high-level as results obtained by an unlabelled semantic segmentation, or mid-level components from an over-segmentation. 
In general, active learning iterates between automated labeling and human input for label rectification~\cite{Yi2016AL}. Compared to conventional learned models with full autonomy, such a {\em human-in-the-loop\/} approach provides a viable option to achieve \rzz{close to error-free} part labeling on test sets, barring human errors\footnote{\rzz{During testing, accuracy is measured against GT labeling which has been produced by humans. To achieve full accuracy, two criteria must be met: (a) the {\em testing\/} human labeler's judgement of what is the correct labeling agrees with the GT; (b) the labeler does not make any error, e.g., due to carelessness or time pressure. We regard both types of violations as ``human errors." One may also refer to the former as an inconsistency between human labeling, which is likely to occur near fuzzy part boundaries.}}.

As shown in \cref{fig:pipeline}, our interactive labeling tool iteratively verifies or modifies part labels predicted by a deep neural network, with human feedback continually improving the network prediction. Specifically, our system consists of three modules for label proposal, verification, and modification. Label proposals are produced by dynamic graph CNN (DGCNN)~\cite{DGCNN}, with the resulting label probabilities dictating whether to pass the predicted labels to the verification or modification module. Both human-verified and corrected labels are passed back to the label proposal network to fine-tune it. The iteration terminates once all part labels have passed human verification.

As the key criterion for success with active learning is the minimization of human effort, we design and incorporate two novel features into our labeling tool:

\begin{itemize}
    \item {\em Hierarchical\/} labeling: We organize all the part labels in a hierarchical tree structure, which guides our prediction-verification-modification process so that labels further down in the hierarchy are dealt with only {\em after\/} their parent labels have been fully verified. \rz{This leads to labeling efficiency, since our labeling process is {\em coarse-to-fine\/}, with fewer labels to process in the coarser levels of the hierarchy.} In addition, with both human users and the prediction network only needing to deal with labels at the same level of the hierarchy rather than across all levels, there is less load on the users to reduce possible errors and the prediction accuracy improves, which reduces the amount of label modifications, which are the most expensive task.
    \item Part {\em symmetry\/}: Since symmetric parts receive the same label, we employ detected symmetries to constrain and facilitate both label verification and modification.
\end{itemize} 

Our hierarchical active learning tool is coined HAL3D. We adopt the hierarchical labels from PartNet~\cite{mo2019partnet} and train and test our labeling tool on both PartNet, which comes with semantic part segmentations, and Amazon-Berkeley Objects (ABO). ABO is a recently published dataset of high-quality, artist-created 3D models of real products sold online. These models are composed of build-aware connected components, However, since some of these components are too coarse, we first obtain an over-segmentation via convex decomposition~\cite{Wei2022} and then apply HAL3D.

We evaluate the core components of HAL3D, \ie, the label proposal network being fine-tuned by human feedback, 
and both hierarchical and symmetry-aware active labeling, by comparing to baselines and performing an ablation.
The results demonstrate clear efficiency of HAL3D over labeling tools without active learning and non-hierarchical designs. Overall, HAL3D achieves 80\% speed-up over manual fine-grained labeling, on the PartNet test set.
\rzz{A supplementary video shows our interactive tool at work. We will release the finely labeled ABO models to serve the community pending permission from the dataset owner.}
\section{Related Work}
\label{sec:related}

Most approaches to semantic 3D shape segmentation and labeling have been designed to reason about coarse- or high-level structures and to target fully autonomy. Considerably less work has been devoted to fine-grained segmentation and labeling or human-in-the-loop approaches.


\mysubsubsection{Coarse 3D shape segmentation and labeling}
Since the seminal work on learning 3D mesh segmentation and labeling~\cite{Kalogerakis2020} in 2010, many learning methods have been proposed~\cite{DLSegSurvey}.
Prominent supervised methods such as PointNet~\cite{PointNet}, PointNet++~\cite{PointNet++}, and DGCNN~\cite{DGCNN} perform feature learning over point clouds, while MeshCNN~\cite{MeshCNN} develops convolution and pooling layers that operate on mesh edges. These methods, among many more, produce both parts and part labels, but they have only been trained and tested on datasets with coarse segmentations, such as the ShapeNet part~\cite{Yi2016AL} and the COSEG datasets~\cite{Wang2012AL}; most object categories therein were labeled with only 2-5 parts.

High-level parts, e.g., the legs, seats, and backs of chairs, typically possess sufficient structural consistency (e.g., in terms of relative positioning), hence predictability, to facilitate supervised learning. Most unsupervised or weakly supervised approaches also leverage such consistencies. In particular, earlier methods for co-segmentation, i.e., segmenting a {\em set\/} of related shapes altogether, all rely on alignment or clustering, either in the spatial domain~\cite{Golovinskiy2009} or feature spaces~\cite{Sidi2011,Hu2012}. More recently, resorting to DL, AdaCoseg~\cite{AdaCoseg} optimizes a group consistency loss defined by matrix ranks. BAE-Net~\cite{BAE-Net} and RIM-Net~\cite{RIM-Net} obtain co-segmentations by learning branched neural fields.


\mysubsubsection{Fine-grained segmentation}
Fine-level parts are important in many real-world applications since, whether large or small, each of them has a function to serve and often a motion to carry out. However, the structure consistency possessed by major semantic groups is mostly lost among fine parts. Such parts could be obtained via shape decomposition based on geometric properties such as convexity~\cite{vanKaick2014,Wei2022}, cylindricity~\cite{Zhou2015}, or through primitive fitting using cuboids~\cite{Yang2021,Sun2019} or quadrics~\cite{Paschalidou2019,CAPRI-Net}, deformed spheres~\cite{paschalidou2021neural, kawana2020neural}. However, the resulting segmentations are neither semantic nor labeled.

Most learning-based solutions are unsupervised or self-supervised, mainly because annotating fine-grained parts is an extremely tedious task. \fg{Li et al.~\cite{Yi17} learns part labeling from noisy online part graphs annotated by human artists. Yu et al.~\cite{yu2019partnet} trains a recursive
part decomposition network for category-specific, fine-grained, hierarchical shape segmentation.} Wang et al.~\cite{wang2021fgseg} and SHRED~\cite{jones2022shred} develop deep clustering to learn part priors from fine-grained segmentation without part labels, where no semantic labels are produced on test shapes.  Sharma et al.~\cite{MvDeCor2022} perform projective shape analysis~\cite{Wang2013PSA}, where a fine-grained 3D segmentation is obtained by aggregating segmented multi-view 2D images. 
Both of these methods operate fully automatically after training, but their performance falls far short of 100\% accuracy: 32.6\% mean IoU score achieved by~\cite{MvDeCor2022} on PartNet~\cite{mo2019partnet} with fine-grained (Level-3) parts (part counts ranging from 4 to 51), and below 50\% mean IoU by~\cite{wang2021fgseg} on a more challenging dataset with part counts up to 500+.

Instead of operating on low-level shape elements such as points~\cite{wang2021fgseg} and pixels~\cite{MvDeCor2022}, the neurally-guided shape parser~\cite{jones2022parser} (NGSP) is trained to assign fine-grained semantic labels to {\em regions\/} of a 3D shape with improved labeling accuracy: mean IoU score up to 58\% on PartNet. NGSP parses region label proposals generated by a guide neural network using likelihood modules to evaluate the global coherence of each proposal. It is still designed as a fully automatic method, but the authors did suggest the need for human-in-the-loop approaches to ``scale beyond carefully-curated research datasets to ‘in-the-wild’ scenarios."

\mysubsubsection{Hierarchical analysis}
A fine-grained shape understanding is also attainable via hierarchical structural analysis. Some early works perform rule-based reasoning to construct symmetry hierarchies~\cite{Wang2011SYMH} and shape grammars~\cite{Liu2014grammar}, while others work with geometric priors, e.g., convexity~\cite{Attene2008}, for hierarchical shape approximation. Similar to fine-grained segmentation, most learned hierarchical models are also unsupervised or weakly supervised, including co-hierarchical analysis~\cite{vanKaick2013}, hierarchical abstraction~\cite{Sun2019}, recursive implicit fields~\cite{RIM-Net}, and inverse constructed solid geometry (CSG)~\cite{CAPRI-Net,UCSG}. All of these methods aim for full automation and work well only with shallow hierarchies.


\mysubsubsection{Active learning for part labeling}
The classical and core question for active learning~\cite{AL_survey2014,AL_comp_survey2022,AL_survey2020}
is: ``How does one select instances from the underlying data to label, so as to achieve the most effective training for a given level of effort?" Past works have studied uncertainty and diversity for interactive segmentation and labeling of 3D shapes~\cite{Wang2012AL,george2022deep}, medical images~\cite{Top2011}, and large-scale point clouds~\cite{ReDAL}, to name a few. We utilize active learning, \rzz{aiming to approach full labeling accuracy}, regardless whether the test shapes belong to a ``research dataset" or are ``in-the-wild."

Most closely related to our work is the active framework for region annotations by Yi et al.~\cite{Yi2016AL}. Both works share the common goal of producing accurate, human-verified labels. \rzz{On the other hand, their work imposes a fixed human effort budget, while ours does not. Technically, their work relies on a conditional random field to automatically propagate human labels, while our work integrates a DL module~\cite{DGCNN} into an active part labeling tool.} The only prior work that combines deep and active learning for 3D segmentation~\cite{george2022deep} focuses on \rzz{boundary refinement, just as the earlier work~\cite{Wang2012AL} whose learning module is constrained $k$-means clustering.} However, none of these tools were designed to work with fine-grained labels, e.g., \rzz{they still only operate on four part labels for chairs~\cite{george2022deep,Wang2012AL,Yi2016AL,Wu2014}.}

Finally, while our work is the first active learning tool for fine-grained 3D part labeling, hierarchical active learning has been employed in other data domains such as documents. In hierarchical text classification\cite{li2012active, cheng2012active}, text labels can also be organized hierarchically. In comparison, semantic labeling of 3D parts in a shape collection poses various challenges including richer structural variations, part relations, and more costly and delicate user interactions.


\section{Method}
\label{sec:method}

Given a shape test set $S$, where each shape is pre-segmented into parts, our methods assigns a label $l \in L$ to each part of each shape. The set of labels, $L$, is predefined and organized in a tree structure such that coarse labels are at the top, and fine-grained labels at the bottom. The labeling progresses hierarchically top-to-bottom, assigning first coarse labels and refining them at each step, achieving high-accuracy fine-grained labeling results.

At each node, we use an active learning based tool to iteratively engage human interactions into the part labeling process. With this tool, we obtain a \rzz{close to error-free part labeling} on each node on any test set.




\subsection{Hierarchical labeling strategy}
\label{ssec:HL}

\rzz{We organize the part labels into a tree structure based on their semantics, as in PartNet~\cite{mo2019partnet}. Following the same terminology, two types of {\em internal\/} tree nodes} are defined: an “AND” node indicates “what part(s) it has” (e.g., a regular base has legs and a runner), while an “OR” node indicates “what type it is” (e.g., a star- or pedestal-type base). The labeling on “AND” nodes is based on part features, while the labeling on “OR” nodes depends on group features; see \cref{fig:proposal-net}. Each leaf node presents a final part label.


At each node of the tree, we combine a deep neural network with an active learning tool to perform part labeling. Consider a single shape $s_i \in S$ and its set of parts $P_i$. At the root node, the input are all the parts of shape $s_i$, that is $P_{root} = P_i$. After finishing the labeling task at the root node, the set of parts is split into one or more subsets $P_{l} \subseteq P_{root}$ corresponding to the children labels $l$. Then, each subset is input to the respective node for further labeling. By splitting the input set at each node, the size of the input decreases as the depth of the node increases. Finally, the labeling stops when we reach a leaf node.


The complexity of our labeling scheme depends on the structure of the tree. On one extreme, if all labels are direct children of the root node, \rzz{training at the root would be more complex but there is only one node to train. The other extreme is a binary tree, where each node is trained to distinguish between only two labels, but we have to train more nodes, resulting in an increased training time. As we have found the original PartNet trees to have too many internal nodes, 
we modified this initial tree structure using a heuristic rule: we remove internal ``AND'' nodes which contain less than three children and keep all ``OR'' nodes. This way, our tree structure is better balanced for our task to lead to reduced training time.} Examples of our modified tree structures are provided in the supplementary material. 

\subsection{Symmetry-assisted labeling}
Another novel feature to accelerate HAL3D makes use of detectable part symmetries on the test set. Symmetry is ubiquitous in human-made objects and it is common practice for 3D artists to copy-and-paste symmetric or repeated parts when building 3D assets. Symmetry detection has been a well-studied subject in geometry processing~\cite{Mitra2013Sym} and there are a variety of applicable techniques. For simplicity, we simply use sizes of the part oriented bounding boxes as a crude and conservative feature to detect part symmetries.


As symmetric parts are expected to share the same label, we accelerate our active learning in two ways: if a proposal assigns different labels to symmetric parts, we automatically remove it from the ``HC (High confidence)'' set. Then, in the modification module, the user only needs to label one part of the symmetry group and our system automatically propagate it to the other parts. These features improve labeling  efficiency by~20\%; see Section~\ref{ssec:eval_PN}. 

\subsection{Active labeling of fine-grained 3D parts}
\label{ssec:DLAL}

Fine-grained 3D shape segmentation and annotation is challenging because of the wide structural diversity found in small and intricate parts of each shape. This diversity makes it difficult for existing DL methods to achieve high prediction accuracy. On the other side, active learning techniques are commonly used by researchers to improve the performance of the learned classifier with the help of human efforts. We adopt these techniques and, at each node of our hierarchical labeling strategy, we use a deep active learning based method to achieve high part labeling accuracy. Our method comprises three main modules: the DL-based part label proposal network, the part label verification module, and the part label modification module.


Consider a node $l$ in our tree structure, given the subset of parts labeled $l$ of each shape, we iteratively update the part labels until they all pass human verification. At each iteration, we first use a DL-based neural network to make part label proposals based on part label probabilities. Then, the HC proposals are passed to the part label verification module for human verification. Next, we collect the LC, i.e., low-confidence, proposals, which are more likely to be inaccurate, and pass them to the label modification module. Finally, the verified proposals and modified proposals are combined and feed back to the proposal network to fine-tune the network weights. The active learning iteration will stop if the number of left shapes are less than 40. We explain next how we make HC vs.~LC proposals.

\begin{figure}[t!]
     \centering
     \includegraphics[width=0.99\linewidth]{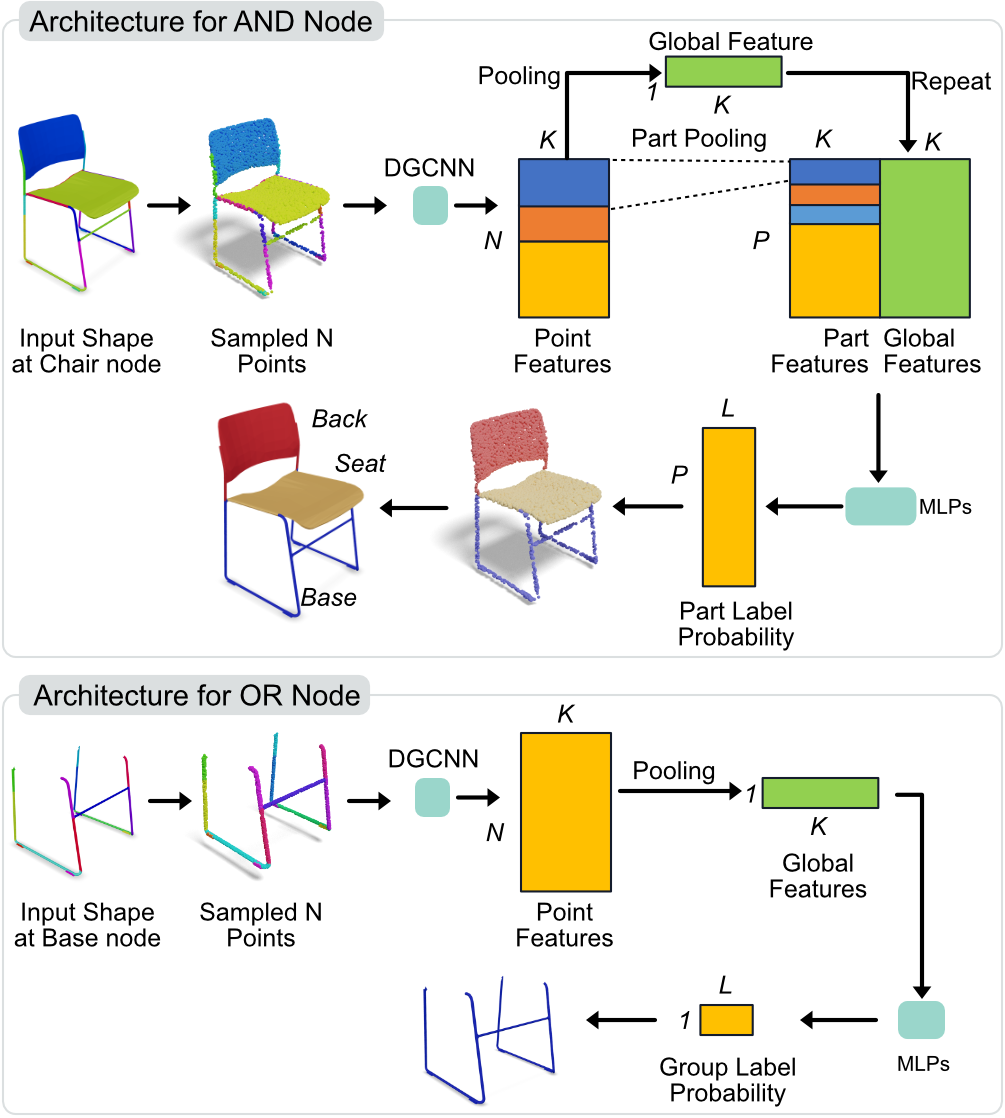}
   \caption{Label proposal architectures. Part label prediction is based on global and local, part-level features at the ``AND" nodes and global features only for the ``OR" nodes. The ``AND" nodes would serve to predict the label for each part and the ``OR" nodes would predict labels for the entire input shape. Notion-wise, $N$ is the number of points, $P$ is the number of parts, $K$ is the feature dimension, and $L$ is the number of labels.}
\label{fig:proposal-net}
\end{figure}
\vspace{-8pt}

\paragraph{Label proposal network.}
%
%
%
Given a set of parts from each shape $s_i \in S$, we  sample $N =$ 8,192 points over the part surfaces and input them into our proposal network. \rzzz{As shown in \cref{fig:proposal-net}, our network uses the edge convolution (EdgeConv) module from DGCNN~\cite{DGCNN} to extract point features from the point clouds. This module consists of five EdgeConv layers whose feature dimensions are 64, 64, 128, 256, and 256, respectively. The point features from all layers are concatenated and passed into a convolution layer. The size of the resulting feature is 512.}

\fg{Since our tree structure has different types of nodes, we apply two different network structures from DGCNN, where the architecture for ``OR" nodes is the same as in the original DGCNN classification network.} At an ``AND" node, differently from DGCNN which is the SOTA method to predict point labels based on {\em point\/} features, we use extra max-pooling operations to aggregate the local {\em part\/} features and global shape features, as inspired by NGSP~\cite{jones2022parser}. The concatenated features are fed into the part classifier, which contains 3 MLP layers to predict part labels. As for an ``OR" node, labels of the input shape are predicted based on the global features only; see \cref{fig:proposal-net}. Since different shapes have different part \rzz{counts}, we set the maximum part count $P_{max} = 150$ during training for batch processing and add zero-padding to the part feature matrix so that all the part feature matrices have the same \rzz{dimension}. More details of our network can be found in the supplementary material. 


\vspace{-8pt}

\paragraph{Label verification.}
Our goal is to get the labels of each shape verified by a human. To reduce verification and training times, we need to decide which shapes are verified first and how many to verify on each iteration.
%
Shapes in the test set that have similar structure to shapes in the train set are more likely to be labeled well.
We can quickly select these well-labeled shapes from the test set and use them to enhance our limited train set, improving the generalization ability of our proposal network. In our method, we first sort the test set based on \fg{average predicted part label probabilities from the proposal network of each shape} and arrange the test set into batches, each containing $B=10$ shapes. Users can mark a shape as well-labeled if they see all parts are labeled correctly, or bad-labeled when one part is wrong. We design a user-friendly interface to achieve this, where users can view and rotate a batch of 3D shapes at the same time. The interface also shows the corresponding label colors for reference. Please see supplementary material for more details about our part label verification interface. 

Given the sorted shapes, we need to decide how many of them to show the user as \fg{HC proposals} before fine-tuning the prediction, to achieve a balance between processing efficiency and utilization of user feedback. To this end, we adopt an adaptive strategy: given the sorted proposals, we stop the verification step when the number of human-verified shapes on a batch is below a threshold, which is set to 4 in our experiments. This simple strategy is effective since early batches tend to have higher labeling accuracy than later batches. \fg{This way, the number of high confidence proposals presented to the user would adapt to the classification purity of the current test set.}


\vspace{-8pt}

\paragraph{Label modification.}
%
%
%
Shapes in the test set whose part structures and geometries differ significantly from those in the train set are less likely to be labeled accurately by the network.
These shapes can be regarded as outliers in the label regression task, but they can be especially valuable during fine-tuning. Human intervention is inevitable in this situation to correct the labels. Specifically, the shapes provided to the label modification module are collected from two sets: \fg{the $Q_1=20$ LC proposals with the lowest mean label probabilities, and those shapes that failed user verification more than $H=2$ times.} We ask users to modify the incorrect labels using our modification interface.

In our part annotation interface, labels are organized hierarchically so users can quickly look up the correct labels. The label-to-color mapping table is also provided to help find the incorrect labels. More details about this interface are provided in the supplementary material.

\fg{Shapes that have been verified and modified by humans are added into the initial training set together to fine-tune our proposal network.}
\rzzz{On the other hand, we skip shapes excluded from the HC and LC proposals. These shapes are combined with shapes that failed to pass the verification less than $H$ times to go to the next iteration for labeling.}



\section{Datasets and Metrics}

\label{sec:data}
\paragraph{Datasets.}

The Stanford PartNet~\cite{mo2019partnet} dataset is a common choice for fine-grained shape analysis~\cite{Wang2018fine,jones2022parser,MvDeCor2022}. We follow NGSP~\cite{jones2022parser} and use the four major categories from PartNet for evaluation: chair, table, lamp, and storage furniture. 
Fine-grained labels in PartNet are organized into three granularity levels, and we adopt the second level in our experiments as in NGSP~\cite{jones2022parser}. We first filter out noisy shapes with erroneous labels. Then from each category, we take 500 shapes and split them into 50/50/400 as train/val/test sets.

\begin{figure}[t!]
     \centering
     \includegraphics[width=0.99\linewidth]{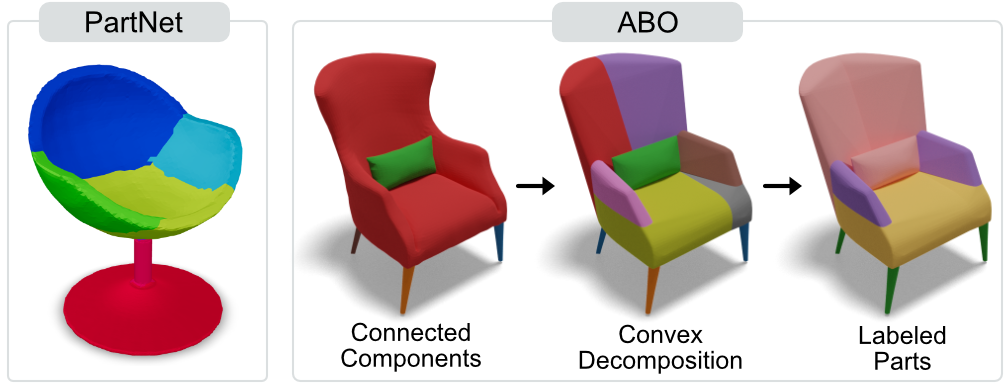}
  \caption{Comparison of pre-segmented parts in PartNet and ABO. ABO shapes were pre-segmented into connected components where each component may correspond to multiple labels, \eg, the back and the arm of the chair are grouped into the same component, as indicated by the red color in the second column. Instead, \rzzz{our} part labeling operates on the convex pieces obtained by a decomposition. \rzzz{In the first three figures, the colors are used to distinguish between different parts, and we can see that the parts in PartNet are manually segmented and exhibiting imperfect boundaries. In the last figure, the colors reflect semantic labels.}}
\label{fig:dataset}
\end{figure}

We also evaluate HAL3D on the recent Amazon Berkeley Objects (ABO) dataset, which is composed of about 8,000 high-quality 3D shapes created by artists from real product images, where each 3D model is a combination of build-aware connected components. Unlike the 3D shapes from PartNet in which each part corresponds to one single semantic label, a build-aware component from ABO may be appropriately associated with multiple labels, as shown in~\cref{fig:dataset}. To alleviate this issue, we apply an approximate convex decomposition~\cite{Wei2022} to further decompose each connected component into a fine-grained, i.e., over-segmented, set of convexes, where each convex piece is deemed to receive one and only one part label. Our active labeling is then applied to the set of convexes. When running the convex decomposition algorithm, we use the default parameters from the provided code, except for the concavity threshold setting, which is set as 0.2 in our experiments.


As GT labels are unavailable in ABO, \rzz{three artists were hired to create them for the decomposed convex components. We built a test set containing 1,800 shapes from five categories: chair, table, lamp, storage furniture, and bed, with 200 for storage and 400 each for the other categories.}

\vspace{-8pt}

\paragraph{Metrics.} 
Active learning systems are typically evaluated from two perspectives: labeling accuracy and the amount of human efforts or involvements. We employ two standard metrics to assess the labeling quality: (1) part label accuracy measuring how many parts are correctly labeled and (2) point cloud segmentation mIoU depicting how much of the shape surface area is correctly labeled.

The amount of human efforts is measured by the consumed time in the steps of label verification and modification. Inspired from~\cite{Yi2016AL}, we firstly give artists the same training courses and tools, and conduct user studies on PartNet. We report the total recorded labeling time in \cref{tab:prop_compare}. We also collect the operations they used in the user study and model the total human interaction time as a function of the number of processed parts and shapes:
\begin{equation}
    t = T^p_v(P_v, L) + T^f_v(S_v, L) + T^p_v(P^c_m, L)+ T^p_m(P^w_m, L),
    \label{eq:time1}
\end{equation}
where $P_v$ is the number of correctly labeled parts in verification, $S_v$ is the number of incorrectly shapes in verification, $P^c_m$ is the number of correctly labeled parts in modification, $P^w_m$ is the number of incorrectly labeled parts in modification, and finally $L$ is the total number of labels. 

Function (\ref{eq:time1}) is devised based on several observations: a) users need to go through and verify all the part labels to ensure accuracy; b) they can easily pick up failure cases during verification if any part is incorrectly labeled, without checking other parts; and c) users need to check all the parts in each shape to know whether modification is needed.

The \rzzz{formulas and their coefficients in (\ref{eq:time2}) are obtained via} \fg{linear regression using the total operation and labeling times} collected from all annotators in our user study:

\begin{equation}
\begin{aligned}
    T^p_v(P_v, L) &= (0.31+0.03*L)*P_v, \\
    T^f_v(S_v, L) &= (1.47+0.05*L)*S_v, \\
    T^p_v(P^c_m, L) &= (0.31+0.03*L)*P^c_m, \\
    T^p_m(P^w_m, L) &= (5.28+0.23*L)*P^w_m,
    \label{eq:time2}
\end{aligned}
\end{equation}
where $T^p_v$ is the checking time for correctly labeled parts, $T^f_v$ is the verification time for incorrectly labeled shapes, and $T^p_m$ is the time for part label modification.
We use these functions to evaluate the human interaction timing, \rzz{in seconds,} for hyper-parameters analysis on PartNet. \fg{We also report additional machine time cost} \rzzz{from} \fg{training and fine-tuning in the supplementary material.}

\section{Experiments}
\label{sec:exp}


We have implemented our approach in PyTorch. The label proposal network \fg{at each node} is pre-trained with the Adam optimizer on the training set of PartNet for 250 epochs, where the learning rate is 0.001 and it is decreased by 0.8 every 25 epochs. When fine-tuning the network in HAL3D, the learning rate is fixed to 0.0001, \rzzz{the maximum epoch number is 125, and the validation set is used for early stopping}. Our experiments were ran on an NVIDIA Tesla V100 GPU, and the training time varies from 20 minutes to 3 hours according to the number of parts at each node.

\subsection{Competing methods}
We compare HAL3D, our hierarchical active labeling tool, with two automatic part labeling methods,

\vspace{0.1cm}
$\bullet$ {\bf PartNet}~\cite{mo2019partnet}, which employs the PointNet++~\cite{PointNet++} backbone for point-wise feature extraction and labels parts by a voting scheme over points belonging to the same part.

\vspace{3pt}

$\bullet$ {\bf NGSP}~\cite{jones2022parser}, which aggregates point-wise features via PointNet++~\cite{PointNet++} to obtain part features and then leverages a likelihood-aware beam search for optimization; this represents the state of the art for fine-grained labeling.

\vspace{2pt}

And two variants of our labeling approach,

\vspace{2pt}
 
$\bullet$ {\bf Our prop.} is one such variant without using the manual label verification or modification module; it predicts part labels using the label proposal network only. 

\vspace{2pt}

$\bullet$ {\bf DAL3D} is a deep active learning variant of our approach without using the hierarchical labeling strategy. 

\subsection{Evaluation on PartNet}
\label{ssec:eval_PN}

All methods presented in this section are pre-trained and tested on the \fg{same} training and test sets of PartNet.

\begin{table}[t!]
\begin{center}
\caption{Quantitative comparison against the competing labeling methods and variants. In the table, ``AL" indicates whether a method uses active learning. Since PartNet, NGSP, and Our prop.~are automatic methods, their entries are marked with a `-' instead. ``Accu." denotes the part label accuracy, ``mIoU" denotes the mean Intersection-over-Union, and ``Lab.~time" denotes the labeling time (hours in total) for the active learning methods. }
\label{tab:prop_compare}
\begin{tabular}{ l|c|c|c|c} 
\hline
Method & AL & Accu.$\uparrow$ & mIoU$\uparrow$ & Lab. time$\downarrow$ \rzz{(h)} \\
\hline \hline
\multicolumn{5}{c}{Chair} \\
\hline
PartNet~\cite{mo2019partnet} & - & 54.90& 31.04	&-	 \\ 
NGSP~\cite{jones2022parser} & -& 68.98 & 42.90	 &-	\\ 
Our prop. & -&	67.45 &41.34  &-	\\ 
DAL3D  & \checkmark & 89.84 & 72.30   & 5.99 \\ 
HAL3D & \checkmark & \textbf{94.13} & \textbf{84.92}  & \textbf{4.34} \\
\hline
\multicolumn{5}{c}{Table} \\
\hline
PartNet~\cite{mo2019partnet} &-&33.21 & 14.92 	&-	 \\ 
NGSP~\cite{jones2022parser} &-&47.11 & 27.94	  &-	\\ 
Our prop. &-&	42.77 & 23.50  &-	\\ 
DAL3D  &\checkmark& 78.53 & 69.82    &9.14  \\ 
HAL3D &\checkmark& \textbf{84.18} & \textbf{74.65}   & \textbf{4.41} \\
\hline
\multicolumn{5}{c}{Storage furniture} \\
\hline
PartNet~\cite{mo2019partnet} &-& 55.82 & 30.86 	&-	 \\ 
NGSP~\cite{jones2022parser} &-& 70.60 & 45.79	 &-	\\ 
Our prop. &-&	69.82 &42.43   &-	\\ 
DAL3D &\checkmark& 84.57 & 71.99   & 4.81 \\ 
HAL3D  &\checkmark& \textbf{93.09} & \textbf{88.72}  & \textbf{3.87} \\
\hline
\multicolumn{5}{c}{Lamp} \\
\hline
PartNet~\cite{mo2019partnet} & -& 38.28 & 11.02 		&-	 \\ 
NGSP~\cite{jones2022parser} &- & 49.76 & 23.49		  &-	\\ 
Our prop. &- & 44.77 & 19.27   &-	\\ 
DAL3D &\checkmark & 83.14 & 69.35 & 4.52 \\ 
HAL3D &\checkmark & {\bf 90.53} & {\bf 79.51}   & {\bf 2.16}  \\
\hline
\end{tabular}
\end{center}
\end{table}


\begin{figure}[t!]
     \centering
     \includegraphics[width=0.99\linewidth]{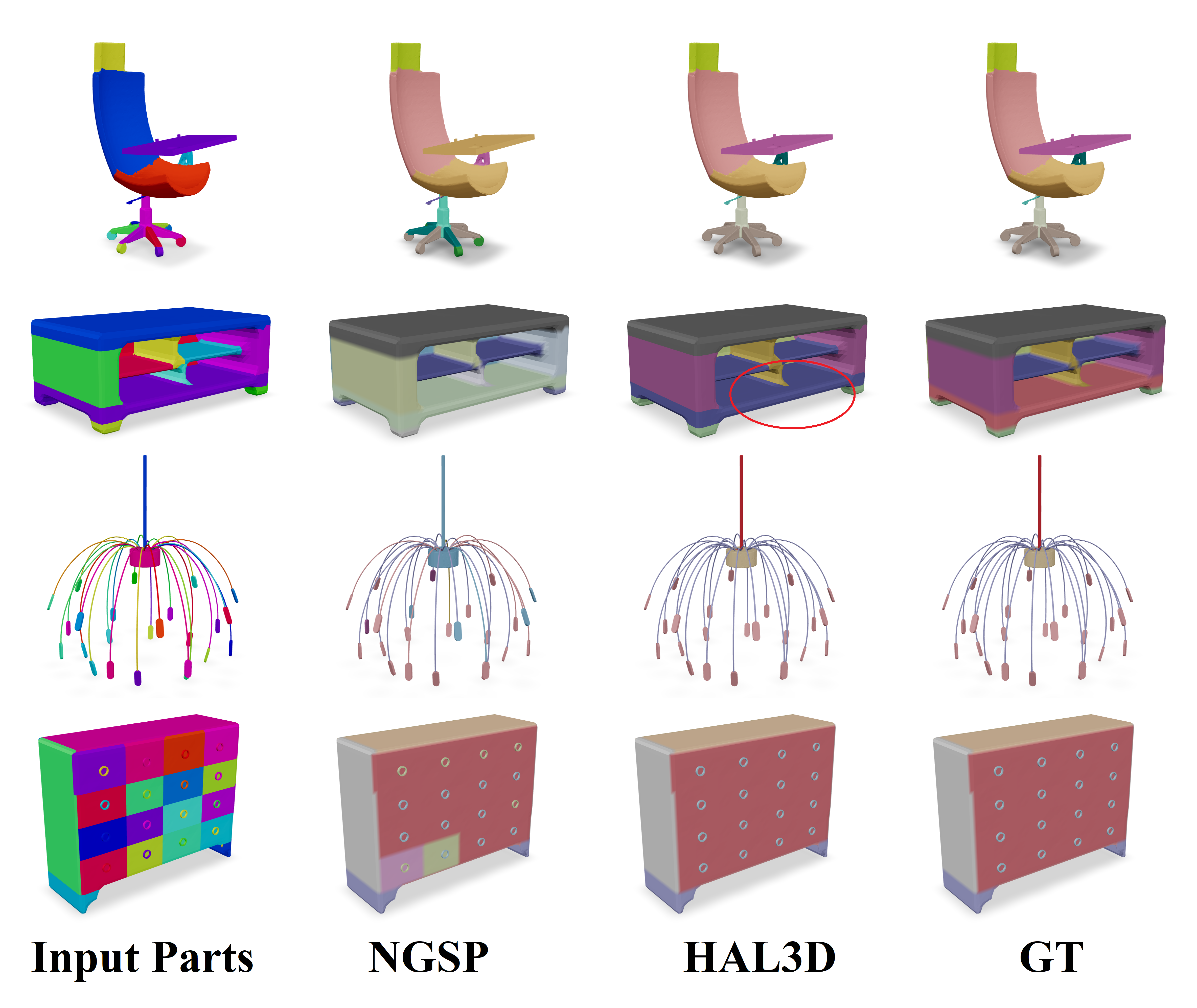}
  \caption{Visual comparison between labeling methods. HAL3D can achieve results approaching GT, barring human errors. The only mis-labeled part by HAL3D here is the bottom panel of the table (red oval), which could be easily labeled as shelf by users.}
\label{fig:visual_compare}
\end{figure}

\vspace{-8pt}


\begin{table}[t!]
\begin{center}
\caption{Ablation on key modules of HAL3D for chair category.}
\label{tab:ablation_new}
\begin{tabular}{c|cccc|cc} 
\hline
Row ID &Prop.&Hier.&Sym.&AL& Lab-T$\downarrow$ & Accu$\uparrow$\\
\hline
2nd &- & - & - & - &  22.05 & 89.16\\
3rd &\checkmark & - & - & - &  8.65 & 88.53\\
4th &\checkmark & \checkmark & \checkmark & - &  6.37 & 93.87\\
5th &\checkmark & - & \checkmark & \checkmark &  5.99 & 89.84\\
6th &\checkmark & \checkmark & - & \checkmark &  5.21 & 93.45\\
7th &\checkmark & \checkmark & \checkmark & \checkmark &  {\bf 4.34} & {\bf 94.13}\\
\hline
\end{tabular}
\end{center}
\vspace{-25pt}
\end{table}

\paragraph{Quantitative comparison.} 
\cref{tab:prop_compare} compares the three automatic methods, where PartNet is inferior to both NGSP and our label proposal network due to its hand-crafted voting mechanism. 
Our simple network achieves comparable accuracy compared to NGSP (difference by 0.11 and 3.49 with respect to mIoU and part accuracy, respectively). However, NGSP suffers from significantly longer running time due to the intensive beam search; see the supplementary for a running time comparison. In contrast, our proposal network is effective and efficient, \ie, capable of quickly producing accurate initial labels to reduce the manual workload in verification and modification for active learning.

\cref{tab:prop_compare} also shows that even with the sophisticated design in NGSP, the highest accuracy attained is only 70\%. Hence the performance of automatic fine-grained labeling still falls far below that of active learning, where the largest and smallest accuracy gap is 40.77\% and 22.49\% between NGSP and HAL3D, respectively. 

HAL3D consistently outperforms our DAL3D variant in all metrics, particularly 2.42 hours faster on average on labeling time, demonstrating the importance of our hierarchical labeling order for reducing manual workload. We also provide qualitative comparison in \cref{fig:visual_compare}.

\begin{figure}[t!]
     \centering
     \includegraphics[width=0.99\linewidth]{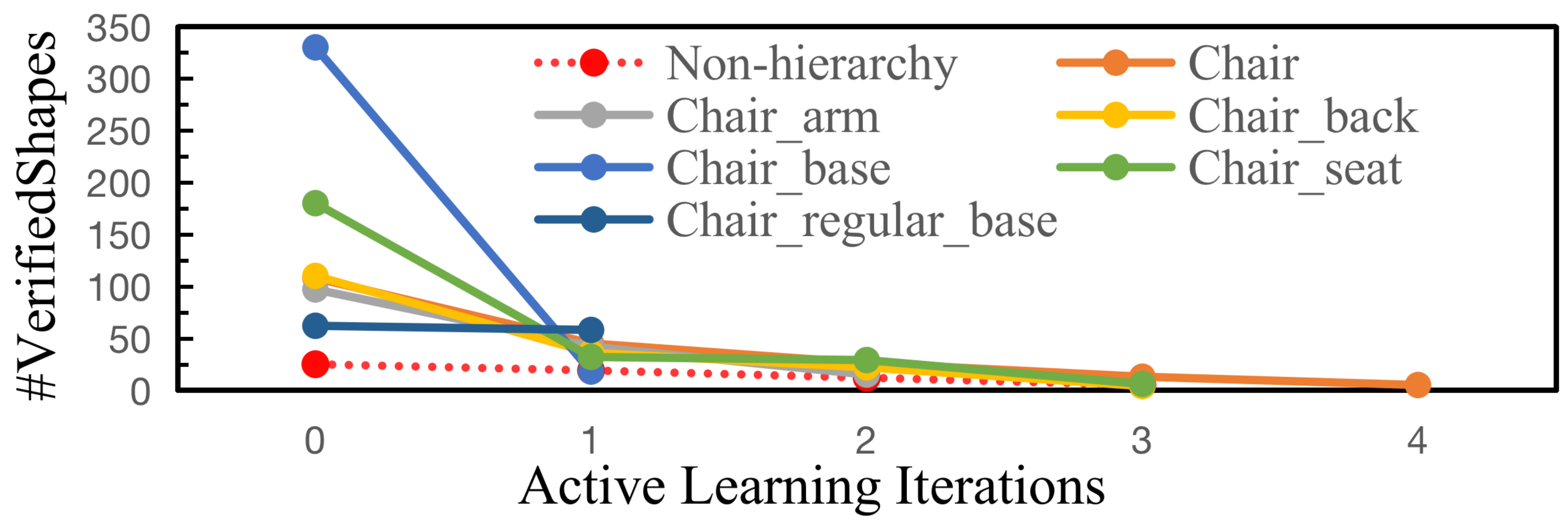}
  \caption{A plot of the number of human-verified shapes (max = 400) at different nodes (only those which existed in more than 100 shapes) of the chair category. The plot shows that with hierarchical active learning, many more shapes pass the verification, hence saving on the costly label modification,
  especially at higher hierarchy levels (i.e., earlier in the labeling iterations).}
\label{fig:AL_itera}
\end{figure}

\vspace{-8pt}

\paragraph{Ablation study.} 
\fg{\cref{tab:ablation_new} verifies the contributions of the main components of HAL3D on minimizing human labeling efforts, a key criterion for success for a human-in-the-loop approach. The second to fifth columns indicate:}

\vspace{2pt}

\noindent $\bullet$ {\bf Prop.} if the label proposal network is used;

\vspace{2pt}

\noindent $\bullet$ {\bf Hier.} if the hierarchical labeling strategy is used;

\vspace{2pt}

\noindent $\bullet$ {\bf Sym.} if the symmetry checking is used.

\vspace{2pt}

\noindent $\bullet$ {\bf AL.} if the active learning is used.

\vspace{2pt}
\rzzz{We compare our full approach against five baselines. The second row is a baseline that directly predicts labels by annotating with our label modification interface. The third row is a baseline that firstly uses the label proposal network and then manually modifies incorrect part labels. The fourth, fifth, and sixth rows represent variants of HAL3D without hierarchical labeling, symmetry checking, and active learning, respectively. The results in \cref{tab:ablation_new} clearly justify the design choices in our method, where our full approach (last row) is the most efficient.}

To further evaluate the effectiveness of our hierarchical labeling, we plot in \cref{fig:AL_itera} the number of shapes which passed human verification over the active learning iterations. As we can see, HAL3D consistently verified more shapes as correct at all tree nodes when comparing to the baseline that had no hierarchical design, e.g., more than 300 shapes were marked as correct in the first iteration in the chair\_base node. Essentially, our hierarchical verification splits the labeling task into smaller and easier tasks compared to the case of non-hierarchical design, which effectively reduces the burden arising from the subsequent and more costly modification step in terms of human effort. 

In the supplementary material, we perform ablations on the four hyper parameters and thresholds employed in our method. In all the experiments, we fix these parameters and set them based only on a 
small dataset. A more careful parameter tuning should lead to better results.

\begin{table}[t!]
\begin{center}
\caption{Inference time comparison between NGSP and our label proposal network. We report average seconds per shape.}
\label{tab:time_compare}
\begin{tabular}{ c|c|c|c|c } 
\hline
Category & Chair & Table & Lamp & Storage \\
\hline
NGSP & 5.410 & 18.01 & 12.69 & 12.87\\ 
\hline
Our prop. & 0.012 & 0.014 & 0.012 & 0.011\\ 
\hline
\end{tabular}
\end{center}
\vspace{-15pt}
\end{table}

\paragraph{Inference time comparison.}

\rzzz{Although NGSP achieved higher prediction accuracy than our proposal network, as shown in \cref{tab:prop_compare}, our network is more efficient and better suited to be incorporated into an active learning framework. \cref{tab:time_compare} compares the inference times between NGSP and our proposal network. NGSP is highly time-consuming due to the use of beam search, whereas our label proposal network takes only 12ms per shape on average.}

\subsection{Evaluation on ABO}

\label{ssec:ABO}

\begin{figure*}[ht!]
\centering
\includegraphics[width=0.99\linewidth]{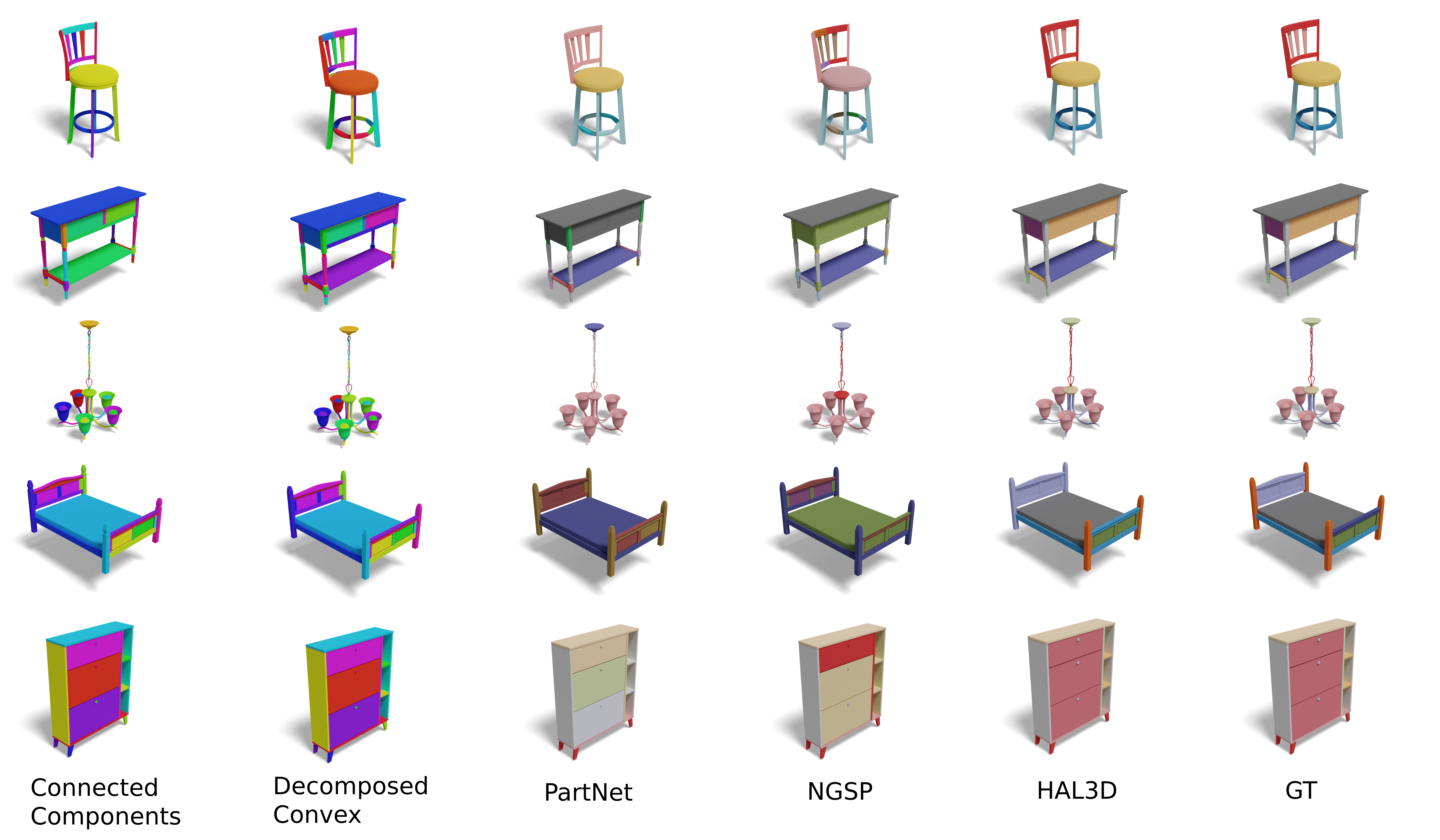}
\caption{\fg{Visual comparison between several part labeling methods on the ABO dataset.}}
\label{fig:abo_compare}
\end{figure*}

We apply HAL3D to label our constructed test set for ABO, where the label proposal network is pre-trained on the training set of PartNet. As mentioned in \cref{sec:data}, shapes in ABO are decomposed into convex shapes and HAL3D assigns a semantic label to each decomposed piece. Three artists were hired to use our HAL3D tool and interface for evaluation. Since GT labels are not provided in ABO, we inspect the labeling results from the artists via human cross validation with the final corrected labeling forming the GT. \rzz{As different artists may report different accuracy and recorded labeling times, we report averages in the paper. The artists worked together with an expert to determine labels for uncertain shapes. \rzzz{The visual results from HAL3D, as shown in \cref{fig:abo_compare} and the supplemental material, were obtained on the test set by taking majority voting from labels provided by different artists.}}

\begin{table}[t!]
\begin{center}
\caption{ Comparison for different methods to pass human verification on test set of ABO. The time is in hours.}

\label{tab:abo_comparison}
\begin{tabular}{ l|c|c|c|c|c } 
\hline
 & Chair & Table & Lamp & Cabinet & Bed \\
\hline \hline
\# shapes & 400 & 400 & 400 & 200 & 400 \\ \hline
\multicolumn{5}{c}{HAL3D} \\ \hline
Time $\downarrow$ & \textbf{6.46} & \textbf{7.41} & \textbf{4.38} &
\textbf{3.34} & \textbf{4.51} \\
Accu. $\uparrow$ & \textbf{93.07} & \textbf{92.55} & \textbf{96.29} & \textbf{94.48} & \textbf{93.71} \\
\hline
\multicolumn{5}{c}{PartNet + modification} \\ \hline
Time $\downarrow$ & 28.25 & 33.94 & 31.53 & 13.35 & 25.02 \\
Accu. $\uparrow$ & 88.96 & 86.51 & 90.91 & 90.24 & 89.37 \\
\hline
\multicolumn{5}{c}{NGSP + modification} \\ \hline
Time $\downarrow$ & 28.30 & 34.32 & 31.56 & 13.22 & 24.90 \\
Accu. $\uparrow$ & 88.77 & 86.40 & 91.37 & 90.56 & 89.93 \\
\hline
\end{tabular}
\end{center}
\vspace{-15pt}
\end{table}

As shown by quantitative results in~\cref{tab:abo_comparison}, HAL3D can accurately label all categories with a 94\% average accuracy using 3-7 hours of manual annotation, demonstrating the effectiveness and efficiency of our method. The reduction on human efforts against manual annotation becomes more significant when comparing against the last four rows in \cref{tab:abo_comparison}, where we use the same manual modification interface to directly correct the PartNet and NGSP results. Please see supplementary material for results on ABO.

\section{Conclusion}
\label{sec:future}

We present HAL3D, the first semantic labeling tool that is designed to operate on fine-grained 3D parts {\em and\/} achieve \rzzz{close to error-free} labeling through full verification by humans, barring human \rzzz{misjudgement}. Our human-in-the-loop approach is based on active learning, combining deep label prediction and human inputs to iteratively fine-tune and improve the network prediction. Extensive experiments demonstrate that human efforts are effectively reduced via hierarchical and symmetry-aware active labeling.

HAL3D is currently implemented with DGCNN~\cite{DGCNN} as the label prediction backbone, but it can be readily replaced by PointNet~\cite{PointNet}, which was adopted by PartNet~\cite{mo2019partnet} and NGSP~\cite{jones2022parser}, or a more performant alternative such as Point Transformer~\cite{PtTrans}.
Aside from symmetry, inter-shape part correspondences can also be utilized to reduce labeling costs, assuming that they can be reliably obtained.
\rzzz{In addition, our tool does not yet allow splitting of under-segmented parts. Since performing such operations by humans is time-consuming, incorporating auto-splitting schemes into HAL3D deserves future consideration.}

We regard our work as only a preliminary step towards deep active learning for fine-grained structural analysis of 3D shapes. A natural focus for future work is to develop a more sophisticated ranking or selection mechanism for label verification and modification to achieve the most effective training for a given level of human effort.

\section*{Acknowledgement}

We thank all the anonymous reviewers and area chairs for their constructive comments and the Amazon artists for performing the human annotations. Thanks also go to Jef-Aram Van Gorp for helpful discussions and other support.

{\small
\bibliographystyle{ieee_fullname}
\bibliography{egbib}
}

\end{document}